\documentclass{article}
\usepackage{ijcai20}

\usepackage{times}
\usepackage{soul}
\usepackage{url}
\usepackage[hidelinks]{hyperref}
\usepackage[utf8]{inputenc}
\usepackage[small]{caption}
\usepackage{graphicx}
\usepackage{amsmath}
\usepackage{amsthm}
\usepackage{booktabs}
\usepackage{algorithm}
\usepackage{algorithmic}
\title{
   ResNetX: a more disordered and deeper network architecture
}

\author{
    Wenfeng Feng$^1$\footnote{Contact Author}\and
    Xin Zhang$^1$\and
    Guangpeng Zhao$^{1}$\\%\And
    \affiliations
    $^1$Henan Polytechnic University, China\\
    \emails
    \{fengwf, zhangxin, gpzhao\}@hpu.edu.cn,
    }
\begin{document}

\maketitle

\begin{abstract}
    Designing efficient network structures has always been the core content of neural network research.
    ResNet and its variants have proved to be efficient in architecture.
    However, how to theoretically character the influence of network structure on performance is still vague.
    With the help of techniques in complex networks, We here provide a natural yet efficient extension to ResNet by folding its backbone chain.
    Our architecture has two structural features when being mapped to directed acyclic graphs:
    First is a higher degree of the disorder compared with ResNet, which let ResNetX explore a larger number of feature maps with different sizes of receptive fields.
    Second is a larger proportion of shorter paths compared to ResNet, which improves the direct flow of information through the entire network.
    Our architecture exposes a new dimension, namely "fold depth", in addition to existing dimensions of depth, width, and cardinality. 
    Our architecture is a natural extension to ResNet, and can be integrated with existing state-of-the-art methods with little effort. Image classification results on CIFAR-10 and CIFAR-100 benchmarks suggested that our new network architecture performs better than ResNet.
\end{abstract}

\section{Introduction}

An artificial neural network is a computing system made up of many simple, \textit{highly interconnected processing elements}, which process information by their dynamic state response to external inputs \cite{caudill_neural_1987}.
How the processing elements are connected is believed to be crucial for the performance of an artificial neural network.
Recent advances in computer vision models also partially confirmed such hyperthesis, e.g., the effectiveness of ResNet \cite{he_deep_2015,he_identity_2016} and DenseNet \cite{huang_densely_2016} and the models in neural architecture search \cite{li_random_2019,pham_efficient_2018,sciuto_evaluating_2019,zoph_neural_2016,zoph_learning_2017} is largely due to how they are connected.

In spite of the architecture of neural networks is critically important, there is still no consistent way to model it till now.
This makes it impossible to theoretically measure the impact of network structure on their performance, and also makes the design of network architecture is based on intuition and more like try and error.
Even if recent models generated by automatically searching in large architecture space are also a kind of try and error method.

On the other hand, the theory of complex networks has been used to model networked systems for decades \cite{newman_networks:_2010}.
If we consider neural networks as networked systems, we can use the theory of complex networks to model neural networks and to characterize the impact of network structure on their performance.
Recently, Testolin et al. \cite{testolin_deep_2018} studied deep belief networks using techniques in the field of complex networks;
Xie et al. \cite{xie_exploring_2019} used three classical random graph models which are theoretical basics of complex networks to generate randomly connected neural network structures.
%Some early works in this direction have been explored \cite{xie_exploring_2019}.

%We here design a new convolutional neural network architecture with the help of techniques in complex networks. Particularly, 
We here first provide a natural yet efficient extension to original residual networks.
By mapping the newly designed convolutional neural network architectures to directed acyclic graphs, we show that they have two structural features in terms of complex networks, that bring the high performance of the model. The first structural feature is they have a less average length of paths and thus a larger number of effective paths, that lead to the more direct flowing of information throughout the entire network. The second structural feature is that those directed acyclic graphs have a high degree of disorder, which means nodes tend to connect to other nodes with different levels, that further improve the multi-scale representation of the model.

\section{Related work}

%-------------------------------------------------------------------------
\subsection{Network architectures}
The exploration of network structures has been a part of neural network research since their initial discovery. Recently, the structure of convolutional neural networks has been explored from their depth \cite{simonyan_very_2014,he_deep_2015,he_identity_2016,huang_densely_2016}, width \cite{zagoruyko_wide_2016}, cardinality \cite{xie_aggregated_2017}, etc.
The building blocks of network architectures also extended from residual blocks \cite{he_deep_2015,he_identity_2016,zagoruyko_wide_2016,xie_aggregated_2017} to many variants of efficient blocks \cite{chollet_xception:_2016,tan_efficientnet:_2019,howard_mobilenets:_2017,sandler_mobilenetv2:_2018,sandler_mobilenetv2:_2018,szegedy_inception-v4_2016}, such as depthwise separable convolutional blocks, etc.

%-------------------------------------------------------------------------
\subsection{Effective paths in neural networks}
Veit et al. \cite{veit_residual_2016} interpreted residual networks as a collection of many paths of differing lengths. 
The gradient magnitude of a path decreases exponentially with the number of blocks it went through in the backward pass.
The total gradient magnitude contributed by paths of each length can be calculated by multiplying the number of paths with that length, and the expected gradient magnitude of the paths with the same length.
Thus most of the total gradient magnitude is contributed by paths of shorter length even though they constitute only a tiny part of all paths through the network.
These shorter paths are called \textit{effective paths} \cite{veit_residual_2016}.
The larger the number of effective paths, the better performance, with other conditions unchanged.

%-------------------------------------------------------------------------
\subsection{Degree of order of DAGs: trophic coherence}
Directed Acyclic Graphs (DAGs) is a representation of partially ordered sets \cite{karrer_random_2009}. The extent to which the nodes of a DAG are organized in levels %( a DAG approach to its state of perfect order)
can be measured by \textit{trophic coherence}, a parameter that is originally defined in food webs and then shown to be closely related to many structural and dynamical aspects of complex systems \cite{johnson_trophic_2014,dominguez-garcia_intervality_2016,klaise_neurons_2016}.

For a directed acyclic graph given by $n \times n$ adjacency matrix $A$, with elements $a_{ij} = 1$ if there is a directed edge from node $i$ to node $j$, and $a_{ij}=0$ if not. The in- and out-degrees of node $i$ are $k_i^{in} = \sum_j a_{ji}$ and $k_i^{out} = \sum_j a_{ij}$, respectively. The first node ($i=1$) can never have ingoing edges, thus $k_1^{in} = 0$. Similarly, the last node ($i=n$) can never have outgoing edges, thus $k_n^{out} = 0$. 

The trophic level $s_i$ of nodes is defined as
\begin{equation}\label{eq:trophiclevel}
   s_i = 1 + \frac{1}{k_i^{in}} \sum_{j} a_{ji} s_j , 
\end{equation}
if $k_i^{in} > 0$, or $s_i=1$ if $k_i^{in}=0$.
In other words, the trophic level of the first node is $s = 1$ by convention, while other nodes are assigned the mean trophic level of their in-neighbors, plus one.
Thus, for any DAG, the trophic level of each node can be easily obtained by solving the linear system of Eq. \ref{eq:trophiclevel}.
Johnson et al. \cite{johnson_trophic_2014} characterize each edge in an network with a trophic distance: $x_{ij} = s_i - s_j$. 
They then consider the distribution of trophic distances over the network, $p(x)$. The homogeneity of $p(x)$ is called trophic coherence:
the more similar the trophic distances of all the edges, the more coherent the network. 
As a measure of coherence, one can simply use the standard deviation of $p(x)$, which is referred to as an incoherence parameter: $q = \sqrt{\langle x^2 \rangle - 1}$.

%-------------------------------------------------------------------------
\subsection{Multi-scale feature representation}
The multi-scale representation ability of convolutional neural networks is achieved and improved by using convolutional layers with different kernel sizes (e.g., InceptionNets \cite{szegedy_inception-v4_2016,szegedy_going_2014,szegedy_rethinking_2015}), by utilizing features with different resolutions \cite{chen_big-little_2019,chen_drop_2019},  and by combining features with different sizes of receptive field \cite{he_deep_2015,he_identity_2016,huang_densely_2016}.
We argue that the degree of disorder of convolutional neural network structures improves their multi-scale representation ability.

%-------------------------------------------------------------------------
\section{ResNetX}
Consider a single image $\mathbf{x}_0$ that is passed through a convolutional network. 
The network comprises $L$ layers, each of which implements a non-linear transformation $F_l(\cdot)$, % and output a image $\mathbf{x}_l$
where $l \ge 1$ indexes the layer. $F_l(\cdot)$ can be a composite function of operations such as Batch Normalization (BN) \cite{ioffe_batch_2015}, rectified linear units (ReLU), Pooling \cite{lecun_gradient-based_1998}, or Convolution (Conv).
We denote the output of the $l^{th}$ layer as $\mathbf{x}_l$.

ResNet \cite{he_deep_2015,he_identity_2016} add a skip-connection that bypasses the non-linear transformations with an identity function:
\begin{equation}\label{eq:resnet}
   \mathbf{x}_l = F_l(\mathbf{x}_{l-1}) + \mathbf{x}_{l-1}
\end{equation}

An advantage of ResNet is that the gradient can flow directly through the identity function (dashed lines in Fig. \ref{fig:diagram}a) from later layers to the earlier layers.

%-------------------------------------------------------------------------
\subsection{ResNetX design}
\begin{figure*}[t]
   \begin{center}
      \includegraphics[width=0.8\linewidth]{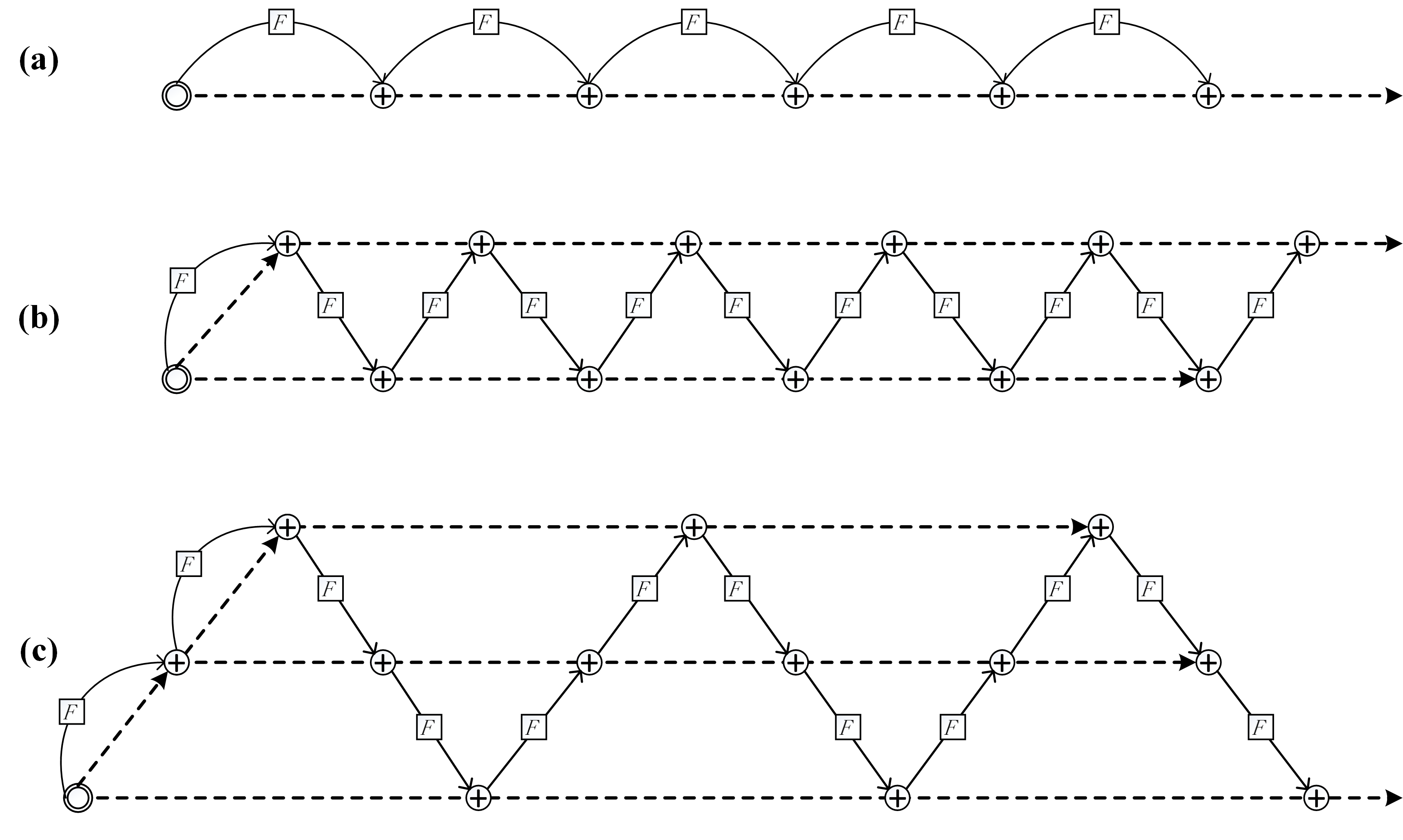}
   \end{center}
      \caption{
         Diagrams of network architectures, (a) for ResNet, (b) for ResNetX when $t=2$, (c) for ResNetX when $t=3$. The double-line circles represent external input image data, the circles with plus signs inside represent summation on all ingoing image data. The dashed lines represent identity functions on image data, while the solid lines (with $F$ on them) represent non-linear transformations on image data.
         }
   \label{fig:diagram}
\end{figure*}

We provide a natural yet efficient extension to ResNet.
Our intuition is simple, we \textbf{fold} the backbone chain (all the non-linear transforms) of ResNet, in order for the direct chain (all the identity functions) to traceback a larger number of previous images with different sizes of receptive fields.
Thus, we introduce a new parameter $t$ to represent the fold depth.
The deeper the fold, the larger the number of previous images of different sizes of receptive fields the model can traceback.
When $t=1$, our model is just the original ResNet.
In order to distinguish our model with ResNet, we name it \textit{ResNetX}, where the character "\textit{X}" at the end is a symbol of the new parameter, i.e. the fold depth. 
Fig. \ref{fig:diagram} illustrated the architectures of original ResNet(\ref{fig:diagram}a), our ResNetX model when $t=2$ (\ref{fig:diagram}b) and $t=3$ (\ref{fig:diagram}c), respectively.
%When we say \textit{ResNet2}, 

Compared with ResNet, our architectures traceback a larger number of previous images that have different sizes of receptive fields, thus promote the fusion of a larger number of images with different receptive fields and improve the multi-scale representation ability. 
Moreover, our architectures increases the number of "direct" chains from one (dashed line in Fig. \ref{fig:diagram}a) in ResNet to two (dashed lines in Fig. \ref{fig:diagram}b), three (dashed lines in Fig. \ref{fig:diagram}c) and more, which decrease the average length of paths through the entire network, increase the number of effective paths, and thus promote the directly propagation of information along with the "direct" chains.
We argue that these two features lead to the effectiveness of our model.

Our model can be formally expressed by the following steps and equations.
First, the output of the current layer $l$, $\mathbf{x}_l$, equal to the summation of the non-linear transformation of the output of the previous layer $F_l(\mathbf{x}_{l-1})$ and the output of layer $l-i$, $\mathbf{x}_{l-i}$:
\begin{equation}\label{eq:resnetx1}
   \mathbf{x}_l = F_l(\mathbf{x}_{l-1}) + \mathbf{x}_{l-i} ,
\end{equation}
The layer difference $i$ is determined by the current layer index $l$ and the fold depth $t$.
When the current layer index is less than the fold depth,  we set $i=1$ like in ResNet, to accumulate enough outputs that could be traced by the later layers, i.e.
\begin{equation}\label{eq:resnetx2}
   i = 1, \;\;\; \textrm{if}  \;\;\; l < t .
\end{equation}
Otherwise, we first divide the current layer index by a number $2(t-1)$ to get the remainder
\begin{equation}\label{eq:resnetx3}
   i_{\cdot} = l \bmod 2(t-1) .
\end{equation}
After that, if the remainder $i_{\cdot}$ is between $[1, t-1]$, the layer difference $i$ equal to $2i_{\cdot}$, i.e.
\begin{equation}\label{eq:resnetx4}
   i = 2i_{\cdot}, \;\;\; \textrm{if}  \;\;\; 1 \le i_{\cdot} \le t-1 .
\end{equation}
Otherwise, we further compute the second remainder
\begin{equation}\label{eq:resnetx5}
   i_{..} = (i_{\cdot} + t-1) \bmod 2(t-1) ,
\end{equation}
and calculate the layer difference $i$ as $2i_{..}$.

In summary, the layer difference $i$ can be computed by the following equation:
\begin{equation}\label{eq:resnetx}
   i = 
   \left\{
      \begin{array}{ll}
      1 & l < t ; \\
      2i_{\cdot} & 1 \le i_{\cdot} \le t-1  ; \\
      2i_{..} & \textrm{else} .
      \end{array}
      \right.
\end{equation}

%-------------------------------------------------------------------------
\subsection{Comparison between ResNetX and ResNet}
In order to compare the architectures of ResNetX and ResNet, we first need to map both of them to directed acyclic graphs. 
The mapping from the architectures of neural networks to general graphs is flexible. We here intentionally chose a simple mapping, i.e. nodes in graphs represent non-linear transformations among data, while edges in graphs represent data flows which send data from one node to another node.
Such mapping separates the impact of network structure on performance from the impact of node operations on performance since all the weights in neural networks are reflected in nodes of graphs.

\begin{figure}[t]
      \begin{center}
         \includegraphics[width=0.9\linewidth]{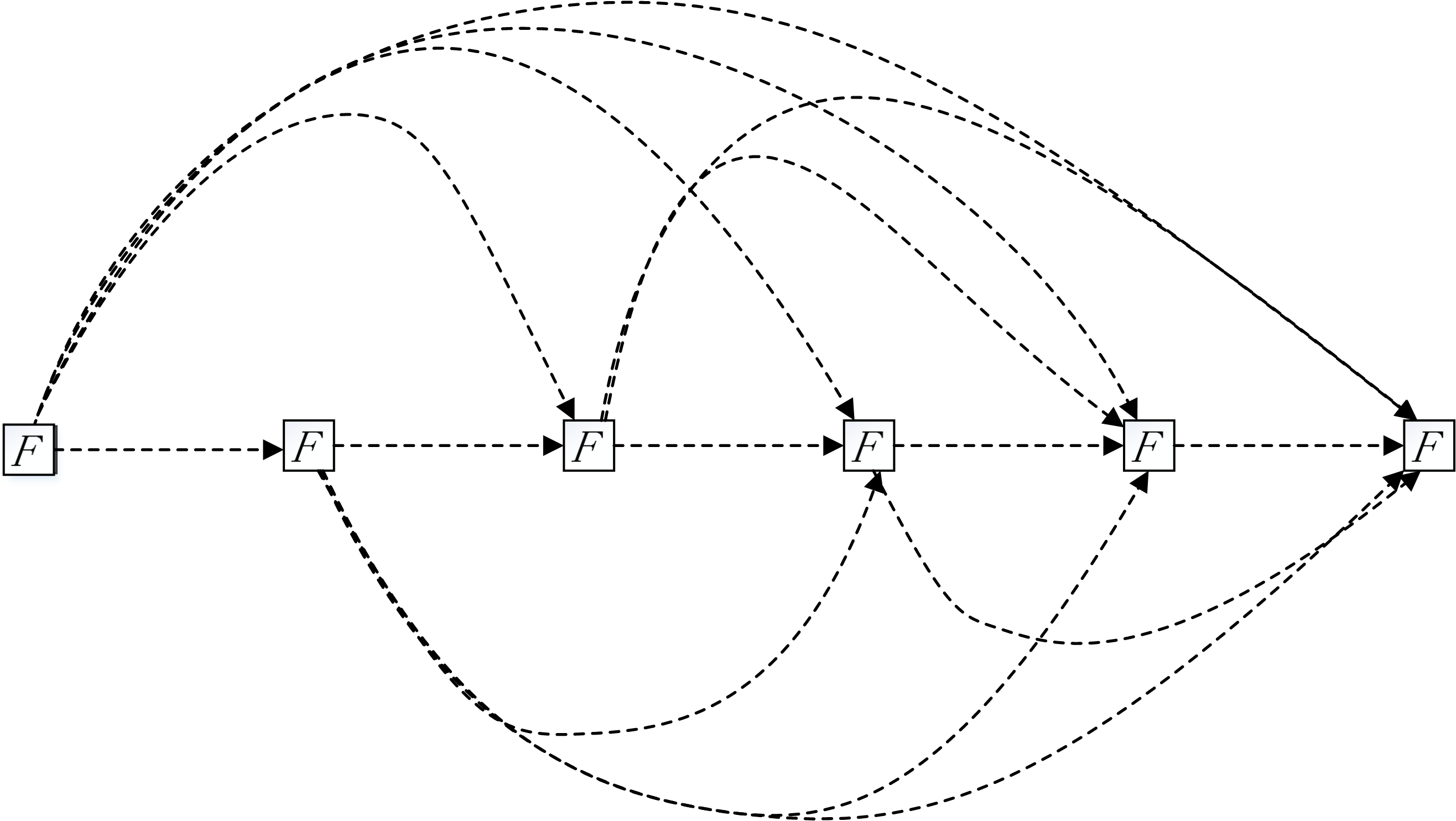}
      \end{center}
      \caption{DAG mapping from ResNet. The square nodes with F inside represent non-linear transformations among data, the dashed lines represent data flows among nodes. }
      \label{fig:diagram2}

      \begin{center}
         \includegraphics[width=0.9\linewidth]{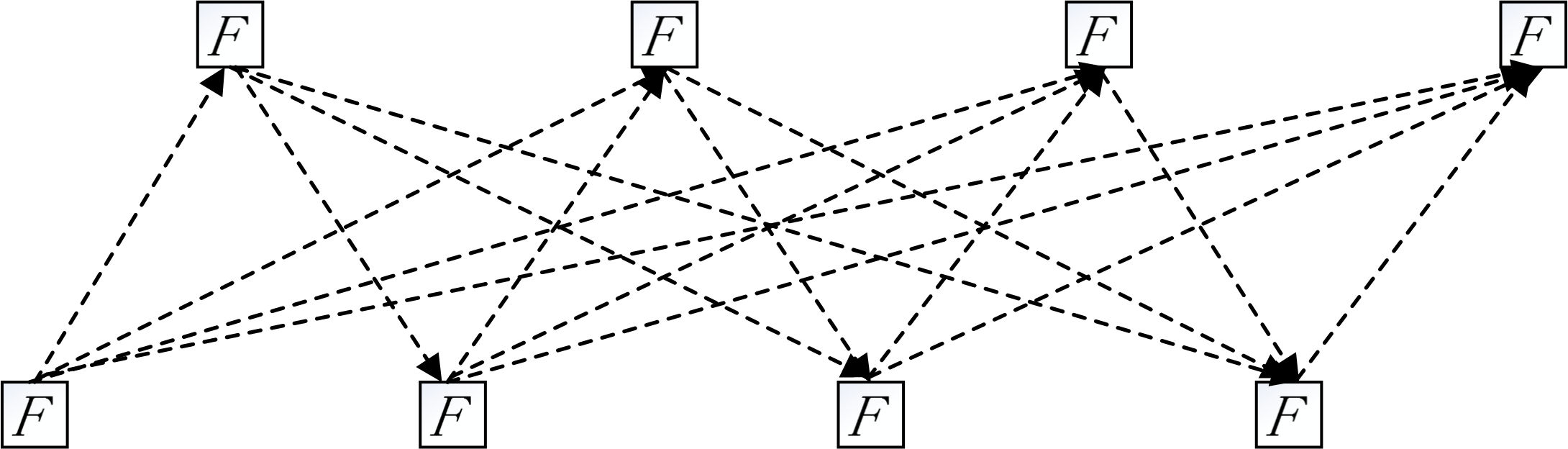}
      \end{center}
      \caption{DAG mapping from ResNetX when $t=2$.}
      \label{fig:diagram3}

      \begin{center}
         \includegraphics[width=0.9\linewidth]{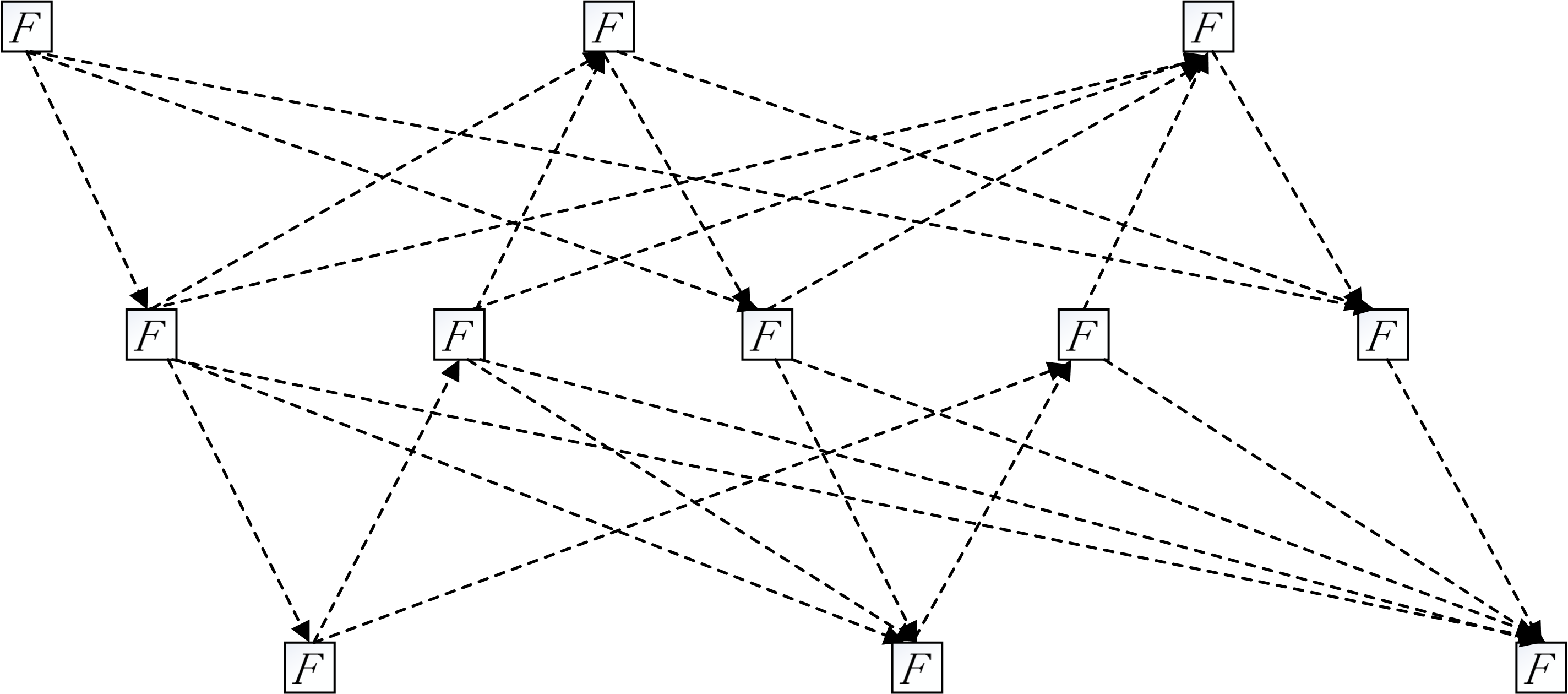}
      \end{center}
         \caption{DAG mapping from ResNetX when $t=3$.}
      \label{fig:diagram4}
\end{figure}

Under the above mapping rule, the architecture of ReNet is mapped to a complete directed acyclic graph (Fig. \ref{fig:diagram2}).
For a complete directed acyclic graph, the distribution of all path lengths from the first node to the last node follows a Binomial distribution, which conforms to results in \cite{veit_residual_2016}.
A complete directed acyclic graph also has a high value of incoherence parameter $q = 0.8523$, which indicates a high degree of disorder.

The architectures of ResNetX are mapped to different directed acyclic graphs according to different values of the fold depth $t$. Fig. \ref{fig:diagram3} and \ref{fig:diagram4} are two examples for $t=2$ and $t=3$ respectively.

\begin{figure}[t]
   \begin{center}
      \includegraphics[width=0.8\linewidth]{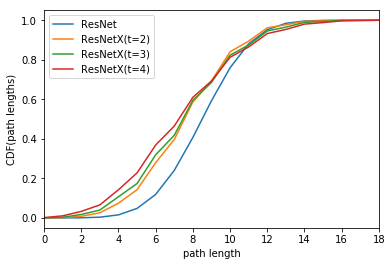}
   \end{center}
      \caption{Comparison of path lengths of ResNet and ResNetX ($n=20$). X axis is the path length, Y axis is the cumulative distribution function (CDF) of path lengths.}
   \label{fig:pathlengths}
\end{figure}

We compared the distribution of path lengths of ResNet and ResNetX in Fig. \ref{fig:pathlengths}. As shown in Fig. \ref{fig:pathlengths}, the proportion of shorter paths of ResNetX are all larger than that of ResNetX, and increase with the fold depth $t$.
We also computed the values of incoherence parameter of ResNetX when $t=2,3,4$ and compare them with the value of incoherence parameter of ResNet. As shown in Tab. \ref{tab:incoherence}, all the values of incoherence parameter of ResNetX are larger than that of ResNetX, and increase with the fold depth $t$.

The comparison of path lengths and incoherence parameter between ResNetX and ResNet show that ResNetX have a larger proportion of shorter paths and a higher degree of disorder than ResNet, and we argue that two features bring better performance of ResNetX. 

\begin{table} 
   \begin{center}
   \begin{tabular}{|l|c|}
   \hline
   Model &  Incoherence parameter ($q$)\\
   \hline\hline
   ResNet & 0.8523 \\  %812059757446
   ResNetX ($t=2$) & 0.8904 \\  % 586637336326
   ResNetX ($t=3$) & 0.8950\\  % 274892513621
   ResNetX ($t=4$) & 0.9124\\  % 149792463323
   \hline
   \end{tabular}
   \end{center}
   \caption{Comparison of incoherence parameters of ResNet and ResNetX.} \label{tab:incoherence}
   \end{table}
   
%-------------------------------------------------------------------------
\section{Experiments}

Limited to experimental conditions, we don't have computing resources to train large-scale data sets.
We have to plan carefully to save very limited computing resources.
Thus, we only consider parameters that are critical for the comparison between ResNetX and ResNet, and keep all other parameters constant. 
Since ResNetX only changes the connecting way of residual connections among earlier and later layers in ResNet, and change nothing inside the layers, it should mainly change the influence of network depth on performance and is orthogonal to other aspects of architecture.
Therefore, we keep all other parameters constant, and only change network depth to evaluate its effect on performance.

We evaluate ResNetX on classification task on CIFAR-10, CIFAR-100 datasets and compare with ResNet.
We choose the basic building block of ResNet and the depthwise separable convolutional block in xception network \cite{chollet_xception:_2016} as the building block of ResNetX, respectively.

%-------------------------------------------------------------------------
\subsection{Implementation details}
Our focus is on the behaviors of extremely deep networks, so we use simple architectures following the style of ResNet-110 \cite{he_deep_2015}.
The network inputs are 32*32 images. The first stem layer is a convolution-bn block. 
Then 4 stages are followed, each stage include $n$ blocks, the number of channels of all stages are set to 32.
The first stage don't down-sample, the other three stages down-sample by maxpool operations.
The network ends with a global average pooling, a 10-way or 100-way fully-connected layer, and softmax. 
The blocks can be the bottleneck block in ResNet or xception block.
The connections among blocks are connected according to the architectures of ResNetX or ResNet.

We implement ResNetX using the Pytorch framework, and evaluate it using the fastai library.
We use \texttt{Learner} class and its \texttt{fit\_one\_cycle} function in fastai library to train both ResNetX and RetNet.
The Adam optimization method and the "1cycle" learning rate policy \cite{smith_super-convergence:_2017} are used.
Momentum of Adam are set to [0.95, 0.85], weight decay is set to 0.01, min-batch size is set to 128, learning rate is set to 0.02, for all situations. 
To save limited computing resources, we run 3 times, each time 5 epochs, for each combination of parameters.
The median accuracy of 3 runs are reported to reduce the impacts of random variations. 
Obviously, we can not output the state-of-the-art results, our goal is to evaluate the relative performance improvements of ResNetX relative to ResNet.

%--------------------------------------------------------------------------
\subsection{DataSets}
The two CIFAR datasets consist of colored natural images with 32*32 pixels. CIFAR-10 consists of images drawn from 10 and CIFAR-100 from 100 classes. The training and test sets contain 50,000 and 10,000 images respectively. We follow the simple data augmentation in \cite{huang_densely_2016} for training: 4 pixels are padded on each side, and a 32*32 crop is randomly sampled from the padded image or its horizontal flip.
For preprocessing, we normalize the data using the channel means and standard deviations. 

%-------------------------------------------------------------------------
\subsection{Results}
For CIFAR-10, blocks per stage $n$ are set to \{24, 32, 40, 64\} respectively, fold depth of ResNetX $t$ are set to \{3, 4, 5\} respectively.
Tab. \ref{tab:results-cifar10-xception} and Tab. \ref{tab:results-cifar10-bottleneck} show the results when the basic block is implemented by xception block and bottleneck block, respectively.
The results show that ResNetX increase the classification accuracy by \textbf{5.42\%} if the basic block is xception block, increase the classification accuracy by \textbf{2.33\%} if the basic block is bottleneck block.

For CIFAR-100, blocks per stage $n$ are set to \{24, 32, 40\} respectively, fold depth of ResNetX $t$ are set to \{3, 4, 5\} respectively.
Tab. \ref{tab:results-cifar100-xception} and Tab. \ref{tab:results-cifar100-bottleneck} show the results when the basic block is implemented by xception block and bottleneck block, respectively.
The results show that ResNetX increase the classification accuracy by \textbf{6.59\%} if the basic block is xception block, increase the classification accuracy by \textbf{2.67\%} if the basic block is bottleneck block.

%the results show that the increase of fold depth $t$ tend to increase accuracy improvements, however 

\begin{table} 
   \begin{center}
   \begin{tabular}{|l|l|l|}
   \hline
   Model & Blocks per stage & Accuracy (\%) \\
   \hline\hline
   ResNet & 24 & 79.69 \\  
          & 32 & \textbf{79.93} \\ 
          & 40 & 79.80 \\ 
          & 64 & 79.72 \\
   \hline
   ResNetX ($t=3$) & 24 & 82.98 \\
          & 32 & 83.85 \\
          & 40 & 83.94 \\
          & 64 & \textbf{84.73} \\
   \hline
   ResNetX ($t=4$) & 24 & 83.86 \\
          & 32 & 84.10 \\
          & 40 & 84.12 \\
          & 64 & \textbf{84.97} \\
   \hline
   ResNetX ($t=5$) & 24 & 83.56 \\
          & 32 & 84.23 \\
          & 40 & 84.39 \\
          & 64 & \textbf{85.35} \\
   \hline
   \end{tabular}
   \end{center}
   \caption{Accuracy of ResNet and ResNetX for CIFAR-10 where basic block implemented by xception block. Bold values are best result for each case.}\label{tab:results-cifar10-xception}
   \end{table}

   \begin{table} 
      \begin{center}
      \begin{tabular}{|l|l|l|}
      \hline
      Model & Blocks per stage & Accuracy (\%) \\
      \hline\hline
      ResNet & 24 & 85.62 \\  
             & 32 & 85.53 \\ 
             & 40 & \textbf{85.74} \\ 
             & 64 & 85.39 \\
      \hline
      ResNetX ($t=3$) & 24 & 86.03 \\
             & 32 & 86.83 \\
             & 40 & 87.40 \\
             & 64 & \textbf{88.07} \\
      \hline
      ResNetX ($t=4$) & 24 & 85.92 \\
             & 32 & 86.57 \\
             & 40 & 86.90 \\
             & 64 & \textbf{87.64} \\
      \hline
      ResNetX ($t=5$) & 24 & 85.86 \\
             & 32 & 86.28 \\
             & 40 & 86.45 \\
             & 64 & \textbf{87.16} \\
      \hline
      \end{tabular}
      \end{center}
      \caption{Accuracy of ResNet and ResNetX for CIFAR-10 where basic block implemented by bottleneck block. Bold values are best result for each case.}\label{tab:results-cifar10-bottleneck}
      \end{table}

   \begin{table} 
      \begin{center}
      \begin{tabular}{|l|l|l|}
      \hline
      Model & Blocks per stage & Accuracy (\%) \\
      \hline\hline
      ResNet & 24 & 46.72 \\  
             & 32 & \textbf{47.15} \\ 
             & 40 & 47.10 \\ 
      \hline
      ResNetX ($t=3$) & 24 & 51.76 \\
             & 32 & 52.09 \\
             & 40 & \textbf{52.91} \\
      \hline
      ResNetX ($t=4$) & 24 & 52.50 \\
             & 32 & 53.13 \\
             & 40 & \textbf{53.74} \\
      \hline
      ResNetX ($t=5$) & 24 & 52.14 \\
             & 32 & 52.90 \\
             & 40 & \textbf{53.52} \\
      \hline
      \end{tabular}
      \end{center}
      \caption{Accuracy of ResNet and ResNetX for CIFAR-100 where basic block implemented by xception block. Bold values are best result for each case.}\label{tab:results-cifar100-xception}
      \end{table}
      
      \begin{table} 
         \begin{center}
         \begin{tabular}{|l|l|l|}
         \hline
         Model & Blocks per stage & Accuracy (\%) \\
         \hline\hline
         ResNet & 24 & 54.87 \\  
                & 32 & 55.27 \\ 
                & 40 & \textbf{55.85} \\ 
         \hline
         ResNetX ($t=3$) & 24 & 56.91 \\
                & 32 & 57.83 \\
                & 40 & \textbf{58.30} \\
         \hline
         ResNetX ($t=4$) & 24 & 56.18 \\
                & 32 & 58.13 \\
                & 40 & \textbf{58.52} \\
         \hline
         ResNetX ($t=5$) & 24 & 55.17 \\
                & 32 & 57.55 \\
                & 40 & \textbf{58.10} \\
         \hline
         \end{tabular}
         \end{center}
         \caption{Accuracy of ResNet and ResNetX for CIFAR-10 where basic block implemented by bottleneck block. Bold values are best result for each case.}\label{tab:results-cifar100-bottleneck}
         \end{table}
            
%-------------------------------------------------------------------------
\section{Conclusion and future work}
We present a simple yet efficient architecture, namely ResNetX.
ResNetX have two structural features when being mapped to directed acyclic graphs:
First is a higher degree of disorder compared with ResNet, which let ResNetX to explore a larger number of feature maps with different sizes of receptive fields.
Second is a larger proportion of shorter paths compared with ResNet, which improve the directly flow of information through the entire network.
The ResNetX exposes a new dimension, namely "fold depth", in addition to existing dimensions of depth, width, and cardinality. 
Our ResNetX architecture is a natural extension to ResNet, and can be integrated with existing state-of-the-art methods with little effort. Image classification results on CIFAR-10 and CIFAR-100 benchmarks suggested that our new network architecture performs better than ResNet.

Although preliminary results suggest the effectiveness of our model, we recognize that our experiments are not enough, and the state-of-the-art results still did not be outputted. We will explore more values of parameters and more datasets if we have feasible conditions. The source code of ResNetX can be accessed at https://github.com/keepsimpler/zero, and we also encourage people to conduct more experiments to evaluate its performance.

\bibliographystyle{named}
\bibliography{resnetx}

\end{document}